\documentclass[10pt,twocolumn,letterpaper]{article}

\usepackage[pagenumbers]{cvpr} 

\usepackage{graphicx}
\usepackage{amsmath}
\usepackage{amssymb}
\usepackage{booktabs}
\usepackage{multirow}
\usepackage{enumitem}
\usepackage[pagebackref,breaklinks,colorlinks]{hyperref}

\def\cvprPaperID{0000} 
\def\confName{CVPR}
\def\confYear{2022}

\begin{document}
\title{Rethinking Spatial Invariance of Convolutional Networks for Object Counting}

\author{Zhi-Qi Cheng\textsuperscript{$1$},
Qi Dai\textsuperscript{$2$\thanks{Corresponding authors.}},
Hong Li\textsuperscript{$3$\thanks{Work done during remote research collaboration with CMU.}},
Jingkuan Song\textsuperscript{$4$},
Xiao Wu\textsuperscript{$3$},
Alexander G. Hauptmann\textsuperscript{$1$\footnotemark[1]}\\
\textsuperscript{1}{Carnegie Mellon University}
\textsuperscript{2}{Microsoft Research}
\textsuperscript{3}{Southwest Jiaotong University} \\
\textsuperscript{4}{University of Electronic Science and Technology of China}\\
{\tt \small \{zhiqic,alex\}@cs.cmu.edu, {qid@microsoft.com},}\\
{\tt \small \{hl1997.work,jingkuan.song\}@gmail.com, {wuxiaohk@home.swjtu.edu.cn}}}

\maketitle
\vspace{-2mm}
\begin{abstract}
\vspace{-2mm}
Previous work generally believes that improving the spatial invariance of convolutional networks is the key to object counting.
However, after verifying several mainstream counting networks, we surprisingly found too strict pixel-level spatial invariance would cause overfit noise in the density map generation.
In this paper, we try to use locally connected Gaussian kernels to replace the original convolution filter to estimate the spatial position in the density map.
The purpose of this is to allow the feature extraction process to potentially stimulate the density map generation process to overcome the annotation noise.
Inspired by previous work, we propose a low-rank approximation accompanied with translation invariance to favorably implement the approximation of massive Gaussian convolution.
Our work points a new direction for follow-up research, which should investigate how to properly relax the overly strict pixel-level spatial invariance for object counting.
We evaluate our methods on 4 mainstream object counting networks~(i.e., MCNN, CSRNet, SANet, and ResNet-50).
Extensive experiments were conducted on 7 popular benchmarks for 3 applications~(i.e., crowd, vehicle, and plant counting).
Experimental results show that our methods significantly outperform other state-of-the-art methods and achieve promising learning of the spatial position of objects\footnote{\url{https://github.com/zhiqic/Rethinking-Counting}}.
\end{abstract}

\vspace{-2mm}
\section{Introduction}
\label{sec:intro}
Object counting has been widely studied since it can potentially solve crowd flow monitoring, traffic management, etc.~The previous works~\cite{Chen-chen,CSR-li,Wang-wang} believe that the latchkey to improving the object counting is to improve the spatial invariance of CNNs.~Based on this starting point, more and more networks (such as dilated CNNs~\cite{ADSCNet-bai,AMDCN-deb,STDNet-ma}, deformable CNNs~\cite{DADNet-guo,ADCrowdNet-liu} and multi-column CNNs~\cite{McML-cheng,AMDCN-deb,EPA-yang}) are studied for object counting.

However, this research direction has appeared performance bottlenecks.
We noticed that the counting accuracy had not been significantly improved with further continuously optimizing the network architectures.
Some recent studies~\cite{SSR-change,SPANet-cheng,S3-lin, MNA-wan} also witnessed a lot of noise during density generation and conjecture that this might be the reason for the performance bottleneck.
Although these efforts have made some progress, we are still ignorant of the following questions.~\textit{1)~Is blindly improving spatial invariance valuable for object counting tasks?~2)~How does density noise affect performance?}

\begin{figure}[t]
\footnotesize
  \centering
   \includegraphics[width=0.75\linewidth]{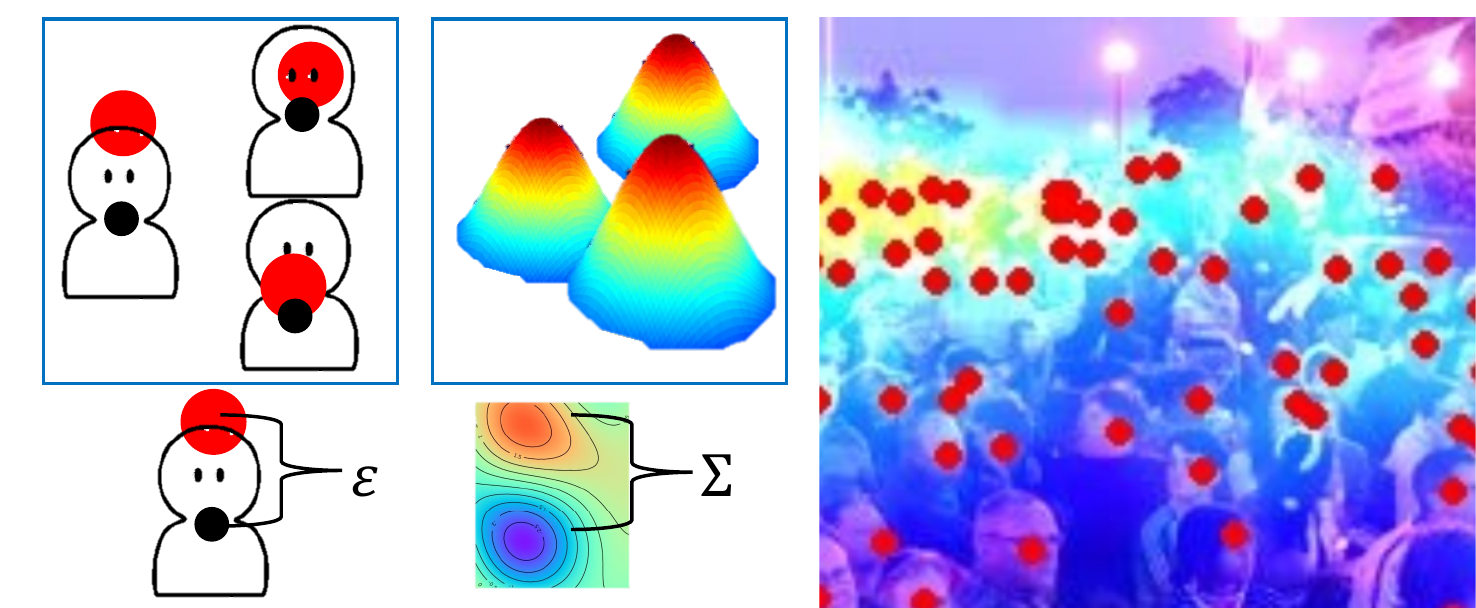}
   \vspace{-2mm}
   \caption{\small
The left shows the idea of density map generation, and the right is an example from SHTech-PartA dataset~\cite{SHTech-zhang}, where the red dot is the annotation in groundtruth, and the black dot is the real center position.
The density map is generated by smoothing the center points with the multi-dimensional Gaussian distribution.
There are two main types of noise:~1)~the error {\large ${\mathbf \epsilon}$} between the true center points and the annotations and~2)~the overlap $\mathbf{\Sigma}$ caused by multiple Gaussian kernels.
[\textbf{Best view in color}].}
    \label{fig:intro}
    \vspace{-6mm}
\end{figure}

Before answering these questions, let’s briefly introduce the generation process of the density map.
Figure~\ref{fig:intro} takes crowd counting as an example.
The density map is generated by smoothing the center point with multiple Gaussian kernels.
This preprocessing converts the discrete counting problem into a continuous density regression, but inevitably brings some noise.
In general, there are two types of noise.~1)~The error between the actual center point and annotation~(i.e., {\large ${\mathbf \epsilon}$} between the red and black dots).~2)~The overlay of Gaussian kernels~(i.e., $\mathbf{\Sigma}$)\footnote{Note that we have some abuse symbols here.}.
More formal mathematical description is in Sec.~\ref{sec:traditional} and \ref{sec:noisy}.

To answer these problems, we have thoroughly verified four mainstream object counting methods~(MCNN~\cite{MCNN-zhang}, CSRNet~\cite{CSR-li}, SANet~\cite{SANet-cao} and ResNet-50~\cite{ResNet-he})~in three different tasks~(crowd, vehicles and plants counting).
Extensive verification experiments reveal that too strict pixel-level spatial invariance will not only cause the large prediction variances, but also overfitting to the noise in the density map as Sec.~\ref{sec:reveal}.
We observed that the existing models~1)~cannot be 
generalized, even impossible within the same crowd counting task and~2)~essentially impossible to learn the actual object position and distribution in the density maps.
In general, these experiments provide the following answers.
\textit{1)~Solely increasing the spatial invariance is not beneficial to object counting tasks.~2)~The pixel-level spatial invariance makes the model easier to overfit the density map noise.}

\begin{figure*}[t]
\footnotesize
  \centering
   \includegraphics[width=0.75\linewidth]{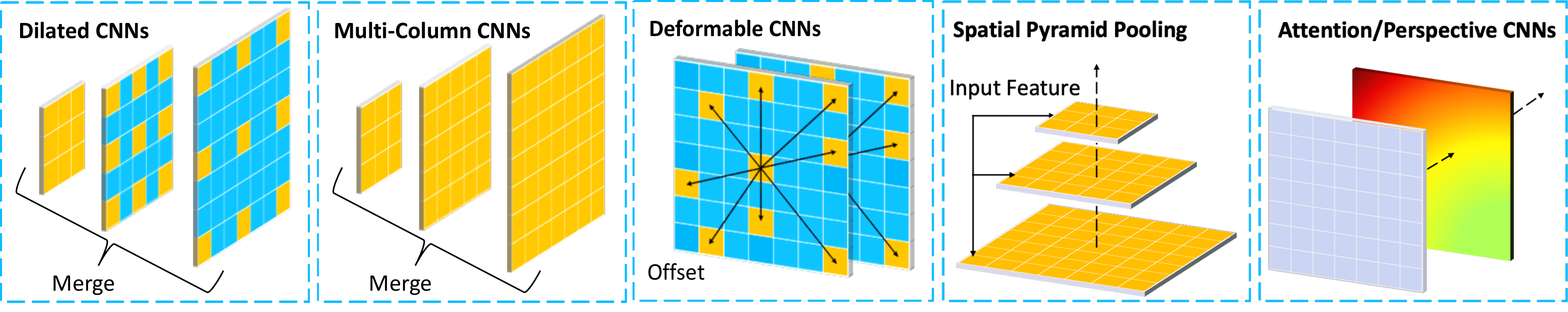}
   \vspace{-2mm}
   \caption{\small 
Overview of research directions of CNNs in object counting.
1)~Dilated CNNs gradually increase the strides of convolution filters to adapt to different sizes.
2)~Multi-Column CNNs utilize different filters to merge features in different branches.
3)~Deformable CNNs optimize the shape of filters to handle multi-scale densities.
4)~Spatial pyramid pooling performs pyramid scaling on input features.
5)~Attention/perspective uses perspective/attention maps through feature extraction.
[\textbf{Best viewed in color}].}
   \label{fig:related-work}
   \vspace{-4mm}
\end{figure*}

To solve these problems, inspired by the previous works \cite{elsayed2020revisiting,Su_2019_CVPR,He_2019_CVPR,tabernik2020spatially}, we try to replace the traditional convolution operation with Gaussian convolution.
The motivation behind this is to mimic the Gaussian-style density generation throughout the whole feature learning, rather than merely generating the final density map.
To a certain extent, this modification is equivalent to a relaxation of the pixel-level spatial invariance.
After the pixel-grid filters are revised with Gaussian kernels, we can jump out of the over-strict pixel-level restrictions.~Fortunately, the experimental result of Sec.~\ref{sec:comparison} proved that this relaxation could allow us to avoid overfitting to the density map noise and promisingly learn the object position and distribution law.

Technically, we propose a novel low-rank approximation to simulate the process of Gaussian-style density map generation during the feature extraction.
Although previous work~\cite{MNA-wan} uses a multivariate Gaussian approximation to optimize the density map in the loss function, it is unclear how to explicitly model this approximation during the convolution process.
Note that the approximation in~\cite{MNA-wan} only imposes the constraints on predicted density maps, while leaving the density estimation unchanged.
In contrast, our approach employs Gaussian convolution to replace standard convolution, where our low-rank approximation uses finite Gaussian kernels (Eq.~\ref{eqn:middle-results}) to approximate the massive Gaussian kernel convolution (Eq.~\ref{eqn:ideal-cnn}). 
It is worth noting that our method concentrates on the density estimation process, while~\cite{MNA-wan} only focuses on the generated density maps.

As shown in Figure~\ref{fig:network architecture}, we replace the standard convolution operation with Gaussian convolution to provide a novel way to generate the density map.
We first propose a \textit{Low-rank Approximation module} to approximate the massive Gaussian convolution.
Specifically, we sample a few Gaussian kernels from the groundtruth density map as input, and then employ Principal Component Analysis (PCA) to select some representative Gaussian kernels.
Through a simple attention mechanism, the correlation between the selected Gaussian kernels is learned, which 
is operated to approximate the massive Gaussian convolution.
Correspondingly, we also propose a \textit{Translation Invariance Module} to accelerate the inference.
On the input side, we adopt the translation invariance to decouple the Gaussian kernel operation to accelerate the convolution operation.
On the output side, we utilize the weights obtained from the low-rank approximation module to accomplish approximation.
Note that all of our implementations are based on CUDA.
It can be seamlessly applied to mainstream CNNs and is end-to-end trainable.
To conclude, our contributions are mainly three folds:
\begin{itemize}[itemsep=2pt,topsep=2pt, parsep=2pt]
\setlength{\topsep}{1mm}
\setlength{\itemsep}{1mm}
\item We reveal that the overly restrictive spatial invariance in object counting is unnecessary or even harmful when facing the noises in the density maps.
\item A low-rank Gaussian convolution is proposed to handle the noises in density map generation.~Equipped with low-rank approximation and translation invariance, we can favorably replace standard convolutions with several Gaussian kernels.
\item Extensive experiments on seven datasets for three counting tasks (i.e.~crowd, vehicle, plant counting) fully demonstrate the effectiveness of our method.
\end{itemize}

\vspace{-2mm}
\section{Related works}
\label{sec:related}
\vspace{-1mm}
We divide the literature into two directions as follows.
\vspace{-2mm}
\subsection{Increase the spatial invariance with CNNs}
\label{subsec:RE:imporve the spatial}
\vspace{-1mm}
Different from traditional manually designed counting detectors~\cite{Chan-chan,Rodriguez-rodriguez,Arteta-arteta,Ma-ma}, existing mainstream methods convert counting problems into density regression~\cite{Chen-chen,Wang-wang,Zhang-zhang,LempitskyCSR-lempitsky}.
The main research direction is to improve the spatial invariance of CNNs.
The mainstream technical routes include Multi-Column CNNs~\cite{MCNN-zhang,EPA-yang, McML-cheng, AMDCN-deb}, Dilation CNNs~\cite{ADSCNet-bai,CSR-li,PFDNet-yan,STDNet-ma,DADNet-guo,AMDCN-deb}, Deformable CNNs~\cite{ADCrowdNet-liu, DADNet-guo}, Residual CNNs~\cite{RRSP-lian,Cascaded-zhao,ResnetCrowd-marsden}, Graph CNNs~\cite{HyGnn-luo}, Attention Mechanism~\cite{ASNet-jiang,RANet-zhang,SOFA-Net-duan,SDANet-miao,ANF-zhang,IA-DNN-sindagi}, Pyramid Pooling~\cite{SPN-chen,CP-CNN-sindagi,huang2020stacked}, and Hierarchy/Hybrid Structures~\cite{HA-CCN-sindagi,HyGnn-luo}.
With the further optimization of parameters and structures, performance bottlenecks have appeared in these approaches, which makes us have to investigate the underlying reasons behind them.

As shown in Figure~\ref{fig:related-work}, we briefly visualized the ideas of these methods.
From the point of view of convolution, the accuracy can be improved by
1)~relaxing the pixel-level spatial invariance (e.g., Dilation/ Deformable CNNs),
2)~fusing more local features (e.g., Multi-Column CNNs and Spatial Pyramid Pooling),
and 3)~exploiting Attention/ Perspective information.
Inspired by this, we utilize a set of low-rank Gaussian kernels with the attention mechanism to relax spatial invariance and fuse local features by replacing standard convolutions.
Here we only offer one solution, and follow-up work can continue to explore how to properly relax the spatial invariance.

\subsection{Dealing with noise in the density map}
\vspace{-1mm}
\label{RE:Deeling with noise}
Similar to our findings, some studies have also shown notable label noise in density maps \cite{URC-xu,CVCS-zhang,MNA-wan,GP-sindagi}.
The mainstream approaches to overcome noise are to propose loss functions~\cite{MFDC-liu,BinLoss-shivapuja,GLoss-wan,SPANet-cheng,BL-ma,SSR-change,S3-lin}, optimize measurement metrics~\cite{P2PNet-song,S3-lin}, update matching rules~\cite{DM-Count-wang,P2PNet-song}, fine-grained noise regions~\cite{SANet-cao,TopoCount-abousamra,AMRNet-liu,SASNet-song}, strengthen regular constraints~\cite{URC-xu,MNA-wan,AMRNet-liu,AutoScale-xu,DUBNet-oh}, combine extra labels~\cite{CVCS-zhang,GP-sindagi,DKPNet-chen,LA-Batch-zhou,NLT-wang}, and optimize training processes~\cite{PSSW-zhao,SANet-cao,TopoCount-abousamra,RGBT-CC-liu}.
Some recent studies have also started to use adversarial~\cite{ASNet-zou,APAM-wu,MS-GAN-zhou,DG-GAN-olmschenk} and reinforcement learning~\cite{LibraNet-liu} to handle noise in the density learning.

In summary, these approaches do not reveal the correlation between the spatial invariance and the noise of density maps.
Most of them only minimize noise by optimizing the loss or regularization term~\cite{BL-ma,CL-idrees,Bayesian-liu,MNA-wan,AutoScale-xu}.
For example, a recent work called AutoScale~\cite{AutoScale-xu} attempts to normalize the densities of different image regions to within a reasonable range.
Our work is inspired by previous work~\cite{MNA-wan}.
Unlike it only focuses on optimizing the loss, our method attempts to modify the convolution operation to overcome noise during the feature learning.

\vspace{-1mm}
\section{Methods}
\label{sec:methods}
\vspace{-1mm}
To better understand our method, we first briefly review the traditional density map generation to reveal the labeling noises of the object counting task.

\vspace{-1mm}
\subsection{Traditional density map generation}
\label{sec:traditional}
\vspace{-1mm}
The recent mainstream approach turns the object counting task into a density regression problem \cite{LempitskyCSR-lempitsky,CCWld-SFCN-wang,CP-CNN-sindagi}.
For $N$ objects of image $\mathcal{I}$, the center points of all objects are labeled as $\left\{\mathbf{\tilde{D}}_1,...,\mathbf{\tilde{D}}_i,...\mathbf{\tilde{D}}_N \right\}$.
The Gaussian kernel can effectively overcome the singularity in the prediction process.
Thus the density of any pixel in an image, $\forall\mathbf{p}_i\in\mathcal{I}$, is generated by multiple Gaussian kernels as,
\begin{align}
\vspace{-4mm}
\label{eqn:density-map-generation}
\footnotesize
y\left( \mathbf{p}_i \right) & =\sum^N_{i=1}\mathcal{N} \left( \mathbf{p}_i;{\mathbf{\tilde{D}}_i},{\beta}\mathbf{I}\right)\\
& =\sum^{N}_{i=1}\frac{1}{2\pi\beta}\exp(-\frac{1}{2}\left\| \mathbf{p}_i- \mathbf{\tilde{D}}_i \right\|^2_{\beta \mathbf{I}}),
\vspace{-4mm}
\end{align}
where {\small$\mathcal{N}(\mathbf{\tilde{D}_i},\beta\mathbf{I})$} is the multivariate Gaussian kernel, the mean {\small$\mathbf{\tilde{D}_i}$ and the covariance $\beta\mathbf{I}$} respectively depict the center point position and shape of the object.
$\beta$ is the variance of the Gaussian kernel and {\small $\left\| \mathbf x \right\|^2_{\beta\mathbf{I}}=\mathbf x^T(\beta\mathbf{I})^{-1}\mathbf x$} is the square Mahalanobis distance.

\begin{figure*}[t]
\small
  \centering
   \includegraphics[width=0.9\linewidth]{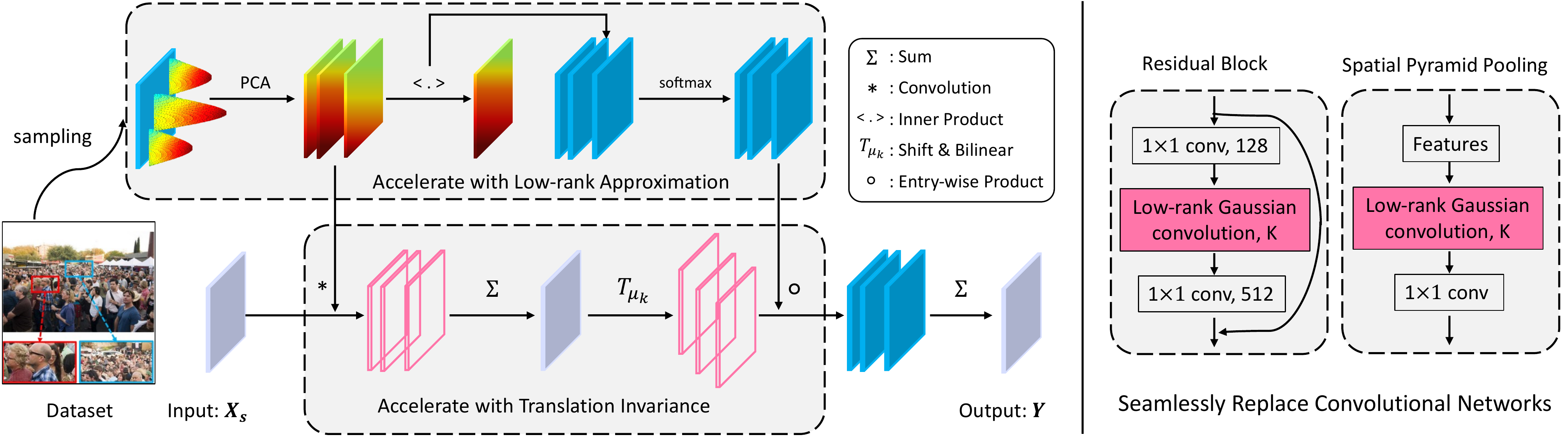}
   \vspace{-2mm}
   \caption{\small 
Illustration of Low-rank Gaussian convolutional layer.
Our proposed layer mainly contains two acceleration modules.
\textit{Low-rank approximation module} has two steps: 1-Principal Component Analysis (PCA) is used to select the Gaussian kernels, 2-Inner product and softmax are used to get the fusion weights.
\textit{Translation Invariance module} also splits the Gaussian kernel operation into two steps: 1-Convolution with Gaussian kernels of zero means, 2-Translation result with other unique means.
Our proposed layer can replace any standard convolutional layer, where the right part is two application examples of residual blocks and pyramid pooling.~[\textbf{Best view in color}].}
   \label{fig:network architecture}
   \vspace{-4mm}
\end{figure*}

\vspace{-1mm}
\subsection{Noise in object counting task}
\label{sec:noisy}
\vspace{-1mm}
However, similar to the previous work~\cite{URC-xu,CVCS-zhang,MNA-wan,GP-sindagi}, we found that there are naturally two kinds of unavoidable noises in density map as Figure~\ref{fig:intro}.
\vspace{-1mm}
\begin{enumerate}[itemsep=2pt,topsep=4pt, parsep=2pt]
\item The error {\large $\mathbf \epsilon$} between the true position of the object {\small{$\mathbf{D_i}$}} and the labeled center point {\small{$\mathbf{\tilde{D}_{i}}$}};
\item The error {$\mathbf{\Sigma}$} between object occlusion and overlapping of multiple Gaussian kernel approximation {\footnotesize {$\sum_{i=1}^{N}\mathcal{N}(\mathbf{p}_i;\mathbf{\tilde{D}}_i,\beta \mathbf{I})$}};
\end{enumerate}

Suppose the labeling error {\large $\mathbf \epsilon$} of the center point position is independent and identically distributed (i.i.d) and also obeys the Gaussian distribution.
Similar to Eq.~\ref{eqn:density-map-generation}, the density map of any pixel {$\forall \mathbf{p}_i\in \mathcal{I}$} with the true center point {\small$\mathbf{D}_i=\mathbf{\tilde{D}}_i-\mathbf{\epsilon}_i$} can also be computed as,
\begin{align}
\vspace{-6mm}
\footnotesize
\label{eqn:truth-density-map}
y\left( \mathbf{p}_i \right)&
=\sum_{i=1}^{N}\mathcal{N} 
\left({\mathbf{p}_i; \mathbf{\tilde{D}}_i-\mathbf{\epsilon}_i},{\beta}\mathbf{I}\right)\\
&=\sum_{i=1}^{N}\mathcal{N} 
\left(\mathbf{q}_i;{\mathbf{\epsilon}_i},{\beta}\mathbf{I}\right),
\vspace{-6mm}
\end{align}
where we have made some equivalent changes to the equations.
Further replacing $\mathbf{p}_i$ with {$\mathbf{q}_i=\mathbf{\tilde{D}}_i-\mathbf{p}_i$}, the density map is still as the combination of the Gaussian distribution $\mathcal{N}(\mathbf{\mu},\mathbf{\Sigma})$.
The values of mean {\large $\mathbf{\mu}$} and variance $\mathbf{\Sigma}$ are respectively estimated as,
\begin{equation}
\vspace{-2mm}
\small
\label{eqn:true-density-expection-mu}
\mathbf{\mu}\approx \mathbb{E}\left[
\sum_{i=1}^{N}
\mathcal{N} \left(\mathbf{\epsilon}_i,\beta\mathbf{I}\right) \right] \approx \sum_i^N \mathbf{\epsilon}_i,
\end{equation}
\begin{equation}
\small
\label{eqn:true-density-expection-sigma}
\mathbf{\Sigma} \approx {\sum_{i=1}^{N} \frac{1}{2\pi \gamma} \mathcal{N}(\mathbf{0},\delta \mathbf{I})-\sum_{i=1}^{N} \mathbf{\mu}_i^2},
\end{equation}
where $\beta,\gamma,\delta$ are the variance parameters of the Gaussian function\footnote{We simply reformulate the parameter by $\gamma=2\beta$ to make a concise expression. The derivation is similar to the previous work~\cite{MNA-wan}, but here the Gaussian distribution is two-dimensional.}.

Although the updated density map still obeys a Gaussian distribution, according to Eq.~\ref{eqn:true-density-expection-mu} and \ref{eqn:true-density-expection-sigma}, the mean {\large $\mathbf{\mu}$} (depicting the center point) and variance $\mathbf{\Sigma}$ (representing shape and occlusion) have more complex forms.
This mathematically sheds light on why strict pixel-level spatial invariance leads to severe overfitting label errors.
As shown in Sec.~\ref{sec:reveal}, some state-of-the-art networks still fail to predict actual occlusion in high-density regions, and overestimate the density in low-density regions.
Obviously, this is due to overfitting to noise, thereby completely ignoring the position and shape of objects.
Below we will present our solution.

\vspace{-1mm}
\subsection{Low-rank Gaussian convolutional layer}
\label{sec:low-rank}
\vspace{-1mm}
Inspired by the previous works~\cite{elsayed2020revisiting,Su_2019_CVPR,He_2019_CVPR,tabernik2020spatially}, we try to replace the standard convolution filters with Gaussian kernels (i.e., propose GauNet).
In this way, the feature extraction can simulate the process of density map generation.~After pixel-grid filters are replaced with Gaussian kernels, we can jump out of the strict pixel-level spatial constraints and learn the density map in a more relaxed spatial manner. 
The modified convolution is as,
\begin{equation}
\label{eqn:ideal-cnn}
\mathbf{Y}_s= \sum_{i=0}^{N}G(\mathbf{\mu}_i,\mathbf{\Sigma}_i)\ast \mathbf{X}_s+\mathbf{b}_s,
\vspace{-1mm}
\end{equation} 
where $\ast$ and $\mathbf{b}_s$ are convolution operation and offsets.
$\mathbf{X}_s$ and $\mathbf{Y}$ are two-dimensional features.
Here we only take the features of channel $s$ as an example.
Since we want to simulate the density map generation, all $N$ Gaussian kernels $G(\mu_i,\Sigma_i)$ have to be used for convolution.
The position and shape of the objects are respectively stipulated by the mean {\large $\mathbf{\mu}_i$} and the variation $\mathbf{\Sigma}_i$.

However, Eq.~\ref{eqn:ideal-cnn} cannot be implemented because it requires 
massive Gaussian convolutions.
Fortunately, previous work~\cite{MNA-wan} uses low-rank Gaussian distributions to approximate the density map.
Inspired by this, we proposed a low-rank approximation module (Sec.~\ref{sec:accelerate-approximation}) to achieve the approximation to Gaussian convolution, and accordingly equipped a translation invariance module (Sec.~\ref{sec:accelerate-invariance}) to accelerate computation.
As shown in Figure ~\ref{fig:network architecture}, we will present these two modules below.

\vspace{-2mm}
\subsubsection{Accelerate with Low-rank approximation}
\label{sec:accelerate-approximation}
\vspace{-2mm}
Low-rank approximation module uses a small number of Gaussian kernels with the low-rank connection to approximate an almost infinite Gaussian convolution (Eq.~\ref{eqn:ideal-cnn}).
It has been proven \cite{MNA-wan} that a density map generated by aggregating $N$ Gaussian kernels ($N$ can be hundreds to thousands\footnote{$N$ is the number of objects in image as shown in Table~\ref{tab:dataset}}) could be approximated by $K$ Gaussian kernels $\left\{ G_1(\mathbf{\Sigma}_1),...,G_K(\mathbf{\Sigma}_K) \right\}$, where $K\ll N$.
Although previous work~\cite{MNA-wan} uses the low-rank approximation to optimize the density map in the loss function, it is still unclear how to approximate the massive Gaussian convolution.

To this end, we try to approximate the infinite Gaussian convolution by learning a few Gaussian kernels, as well as their correlations with an attention mechanism.
During the approximation, a large number of Gaussian kernels are randomly sampled.
After the Principal Component Analysis (PCA), the eigenvectors $\left\{ G(\mathbf{\Sigma}_k) \right\}_{k=1}^K$ corresponding to $K$ non-zero eigenvalues are obtained.
Then we initialize the coefficients of picked $K$ Gaussian kernels as,
\begin{equation}
\mathbf w_k= \left\langle G(\mathbf{\Sigma}_k),G(\mathbf{\Sigma}_I)\right\rangle,
\end{equation}
where $<.>$ is the inner product, and $\mathbf{\Sigma}_I$ represents the identity matrix.
Because we will further decompose the Gaussian kernel to speed up the computation, the mean {\large $\mathbf{\mu}$} of the Gaussian kernels is ignored here.
Finally, we perform normalization operations,
\begin{equation}
\sigma(\mathbf{w}_k)= \frac{\mathrm{exp} \left( \mathbf{w}_k\right)}{\sum_{l=1}^K \mathrm{exp} \left( \mathbf{w}_l\right)},
\end{equation}
where $\mathbf{w}_k$ are also updated during training. 
In addition to fusing the local features, it can also help restrict the spatial information in the gradient back-propagation.

Based on this improvement, the optimized Gaussian convolutional layer is computed as,
\begin{equation}
\label{eqn:middle-results}
\mathbf{Y}_s= 
\sum_{j=0}^{K}(\mathbf{w}_j \circ 
\sum_{i=0}^{K}
(G(\mathbf{\mu}_i,\mathbf{\Sigma}_j)\ast \mathbf{X}_s)
)+\mathbf{b}_s,
\end{equation}
where $\circ$ is the entry-wise product.
We utilize the low-rank Gaussian kernels to complete the approximation process.
Following we will continue to apply the translation invariance module to further optimize our method.

\vspace{-2mm}
\subsubsection{Accelerate with translation invariance}
\label{sec:accelerate-invariance}
\vspace{-1mm}
Translation invariance module aims to decompose the convolution operation between the Gaussian kernel and the input feature map to accelerate the inference.
Accomplishing convolution operations of all Gaussian kernels in Eq.~\ref{eqn:middle-results} requires a lot of computational resources.
Using the translation invariance of Gaussian kernels, the convolution operation between the Gaussian kernel and the input features can be efficiently implemented as,
\begin{align}
G\left( \mathbf{\mu}_k,\mathbf{\Sigma}_k \right)\ast \mathbf{x}=\mathcal{T}_{\mathbf{\mu}_k}\left[ G(\mathbf{0},\mathbf{\Sigma}_k) \right]\ast \mathbf{x}\\
=\mathcal{T}_{\mathbf{\mu}_k}\left[ G(\mathbf{0},\mathbf{\Sigma}_k) \ast \mathbf{x}\right],
\label{eqn:speed-up-gaussian-kernels}
\vspace{-1mm}
\end{align}
where $\mathcal{T}_{\mathbf{\mu}_k}[\mathbf{y}] = g(\mathbf{y}-\mathbf{\mu}_k)$ is the translation of the function $g()$. $G(\mathbf{0},
\mathbf{\Sigma}_k)$ is Gaussian kernels with zero mean.
The benefit of this is that we can ignore the mean of Gaussian kernels in the convolution operation.
Since Eq.~\ref{eqn:speed-up-gaussian-kernels} is only accurate for discrete {\large$\mathbf{\mu}_k$}, we treat the translation function $g()$ as bilinear interpolation in the actual implementation,
\begin{equation}
\label{eqn:bilinear-interpolation}
\tilde{\mathcal{T}}_{\mathbf{\mu}_k}\left[\mathbf{y} \right]=\sum_i\sum_j \mathbf{a}_{ij}\cdot g(\mathbf{y}-\left\lfloor \mathbf{\mu}_k\right\rfloor + \left\lfloor i,j \right\rfloor),
\vspace{-1mm}
\end{equation}
where $\mathbf{a}_{ij}$ are the weights in bilinear interpolation, which allow computing subpixel displacements and can be implemented efficiently in CUDA.

Finally, our proposed  low-rank Gaussian convolutional layer can be computed as,
\begin{equation}
\label{eqn:final-conv}
\mathbf{Y}_s= 
\sum_{k=0}^{K}
(\mathbf{w}_k\circ \sum_{j=0}^{K}(\tilde{\mathcal{T}}_{\mathbf{\mu}_k}[G(\mathbf{\Sigma}_j)\ast \mathbf{X}_s])+\mathbf{b}_s,
\end{equation}
where all implementations are based on CUDA.
Thus our proposed layer can be applied to mainstream CNNs.
In most cases, we replace all the convolutional layers (or
3×3 convolutional layers in all residual and pyramid  pooling blocks)
with our Gaussian convolutional layers.

\textbf{Complexity analysis.}~Theoretically, considering input $\mathbf{X}=[H, W, C_{i}]$ and output $\mathbf{Y}=[H, W, C_{o}]$, supposing $N$ Gaussian kernels are used in density map generation, the complexity of the initial Gaussian convolution~(Eq.~\ref{eqn:ideal-cnn}) is $\mathcal{O}(C_{i}C_{o} H W N k_w k_h)$, where $k_w$, $k_h$ indicate the upper bound of the size of Gaussian kernels.
When utilizing low-rank approximation, the complexity of Eq.~\ref{eqn:middle-results} is $\mathcal{O}(K C_{i} C_{o} H W K k_w k_h)$, where $K$ is the number of the sampled kernels, $K\ll N$.
By further applying translation invariance, the complexity of Eq.~\ref{eqn:final-conv} is $\mathcal{O}(4 K C_{i} C_{o} H W)$, where 4 is related to the bilinear interpolation.
Table~\ref{tab:time} also shows experimental time cost of our method, which demonstrates the effectiveness of two acceleration components. 

\section{Experiments}
\label{sec:experiments}
\subsection{Experiment settings}
\label{sec:experiment-settings}
\vspace{-1mm}
\textbf{Dataset}.
We evaluate our method on three application, i.e., crowd, vehicle, and plant counting.~For crowd counting, five datasets are used for evaluation, including ShanghaiTech (SHTech) PartA and PartB\cite{SHTech-zhang}, UCF\_CC\_50\cite{UCF_CC_50-idrees}, UCF-QNRF\cite{UCF-QNRF-idrees} and JHU-CROWD++\cite{JHU-CROWD-sindagi}.~For vehicle and plant counting, two datasets, i.e., TRANCOS\cite{TRANCOS-guerrero} and MTC\cite{MTC-lu}) are used, respectively.
Table \ref{tab:dataset} gives a summary of these datasets.~

\begin{table}[t]
\centering
\footnotesize
\vspace{-2mm}
\caption{\small
Object counting benchmarks.
[Min, Max] and \#images are the range of objects per image and the number of images.}
\vspace{-2mm}
\begin{tabular}{clrr}
\toprule
\multicolumn{1}{l}{ } & Datasets     & [Min, Max] & \#Images \\
\midrule
\multirow{5}{*}{Crowd}           & SHTech-PartA\cite{SHTech-zhang} & [33, 3,139]      & 482     \\
                                 & UCF\_CC\_50\cite{UCF_CC_50-idrees}  & [94, 4,543]      & 50      \\
                                 & UCF-QNRF\cite{UCF-QNRF-idrees}     & [49, 12,865]     & 1,525   \\
                                 & JHU-CROWD++\cite{JHU-CROWD-sindagi}  & [0, 7,286]      & 4,250   \\
                                 & SHTech-PartB\cite{SHTech-zhang} & [9, 578]        & 716     \\
\midrule                                
Vehicle                          & TRANCOS\cite{TRANCOS-guerrero}      & [9, 107]        & 1,244   \\
\midrule
Plant                            & MTC\cite{MTC-lu}          & [0, 100]        & 361  \\
\bottomrule
\end{tabular}
\label{tab:dataset}
\vspace{-4mm}
\end{table}

\textbf{Baseline Networks}.
We evaluate our method by integrating it with four baselines including MCNN~\cite{MCNN-zhang}, CSRNet\cite{CSR-li}, SANet\cite{SANet-cao}, and ResNet-50\cite{ResNet-he}.
The training procedures follow third-party Github repositories\footnote{https://github.com/gjy3035/C-3-Framework}.
Training details are slightly different from the original paper.
For example, batch processing and other functions are included.
Following previous works~\cite{SPANet-cheng,SDANet-miao,DM-Count-wang}, MCNN and CSRNet are tested on the whole images, while SANet is evaluated on image patches.
Additionally, Mean Absolute Error (MAE) and Mean Square Error (MSE) are used as evaluation measurements.

\vspace{-2mm}
\subsection{Reveal the label noise of object counting}
\label{sec:reveal}
\vspace{-2mm}
We verified the prediction variance on four mainstream object counting methods (i.e.,~MCNN~\cite{MCNN-zhang}, CSRNet~\cite{CSR-li}, SANet \cite{SANet-cao}, and ResNet-50~\cite{ResNet-he}).

\textbf{Large variance in prediction.}
As shown in Figure~\ref{fig:nosiy-analysis}, four object counting methods have a large prediction variance on SHTech-PartA and UCF-QNRF datasets.
Even more surprising is that the variance does not decrease as the performance (spatial invariance) increases.
The results in Figure~\ref{fig:nosiy-analysis} meaningfully reveal its hidden reason, namely that the overly strict pixel-level spatial invariance makes the model severely overfit to the density map noise.

\textbf{Underestimation of high-density areas.}
We performed a similar validation for high-density regions to find out the reasons for the large prediction variance.
In the second column of Figure~\ref{fig:nosiy-analysis}, we noticed that the prediction variance in high-density areas is more severe than the entire image.
The overall statistics prove that the model severely underestimates density in high-density areas.
What is even more surprising is that this variance appears to increase as the performance (spatial invariance) increases.

\textbf{Overestimated in low-density areas.}
Likewise, in the third column, we analyzed the low-density area.~Overall, the variance is reduced compared to high-density areas.
We speculate that fewer Gaussian kernels are in the low-density area, which inherently has lower annotation noise.
Although the variance is slight than the high-density area, the overall variance is still more severe than the entire image.
We guess this is because the high and low-density areas compensate for each other to reduce the variance.

\begin{figure}[t]
\footnotesize
\centering
\includegraphics[width=0.92\linewidth]{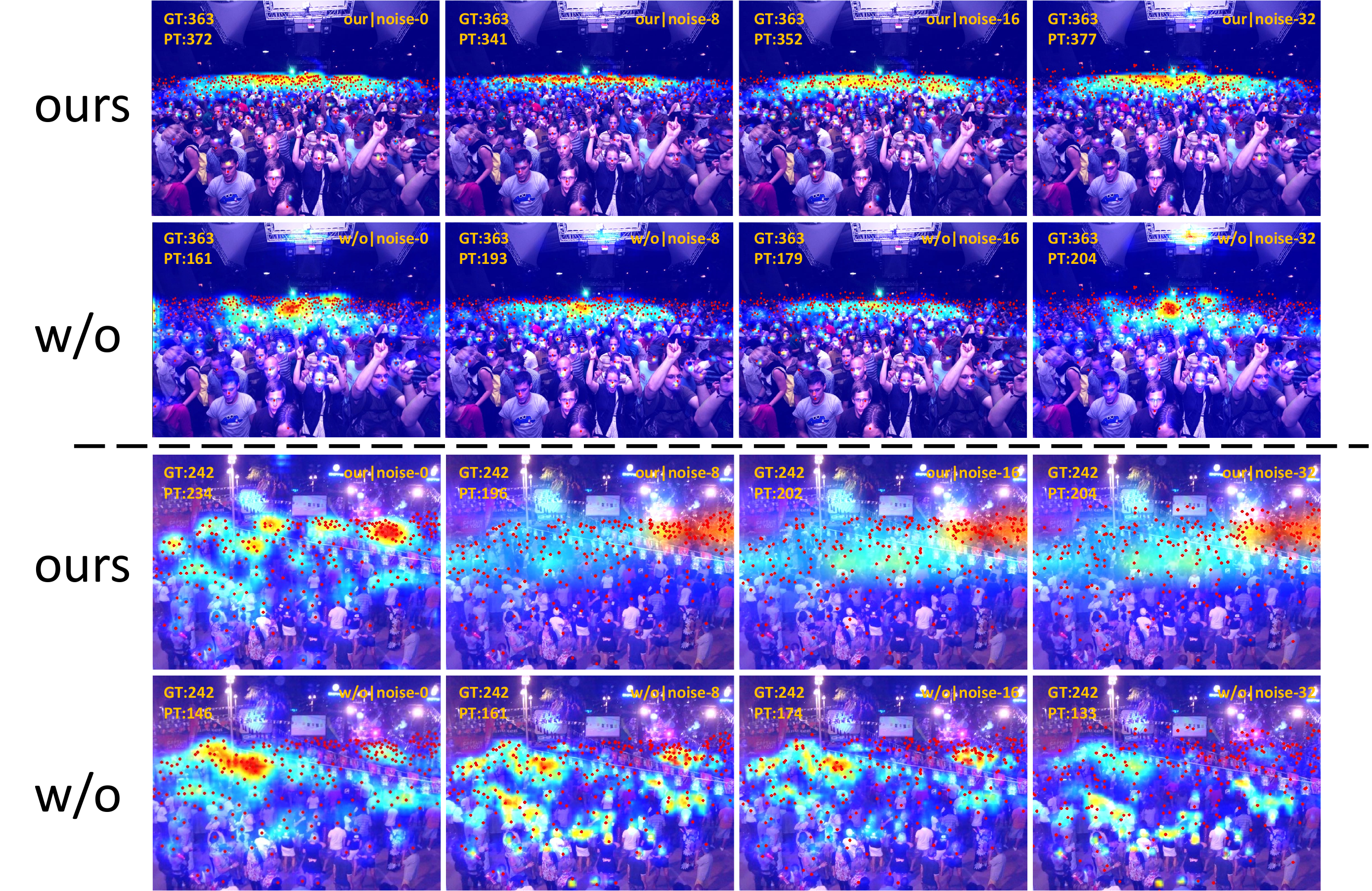}
\vspace{-2mm}
\caption{\small
Visualization of robustness to annotation noise, where red dots are groundtruth annotations.
Here we generate the noisy dataset by randomly moving the annotation points by \{0, 8, 16, 32\} pixels.
Visualization results exhibit the results of two examples with/without our proposed Gaussian convolutional layer.}
\vspace{-2mm}
\label{fig:noisy-visulation}
\end{figure}

\begin{figure}[t]
\footnotesize
 \centering
   \includegraphics[width=1.0\linewidth]{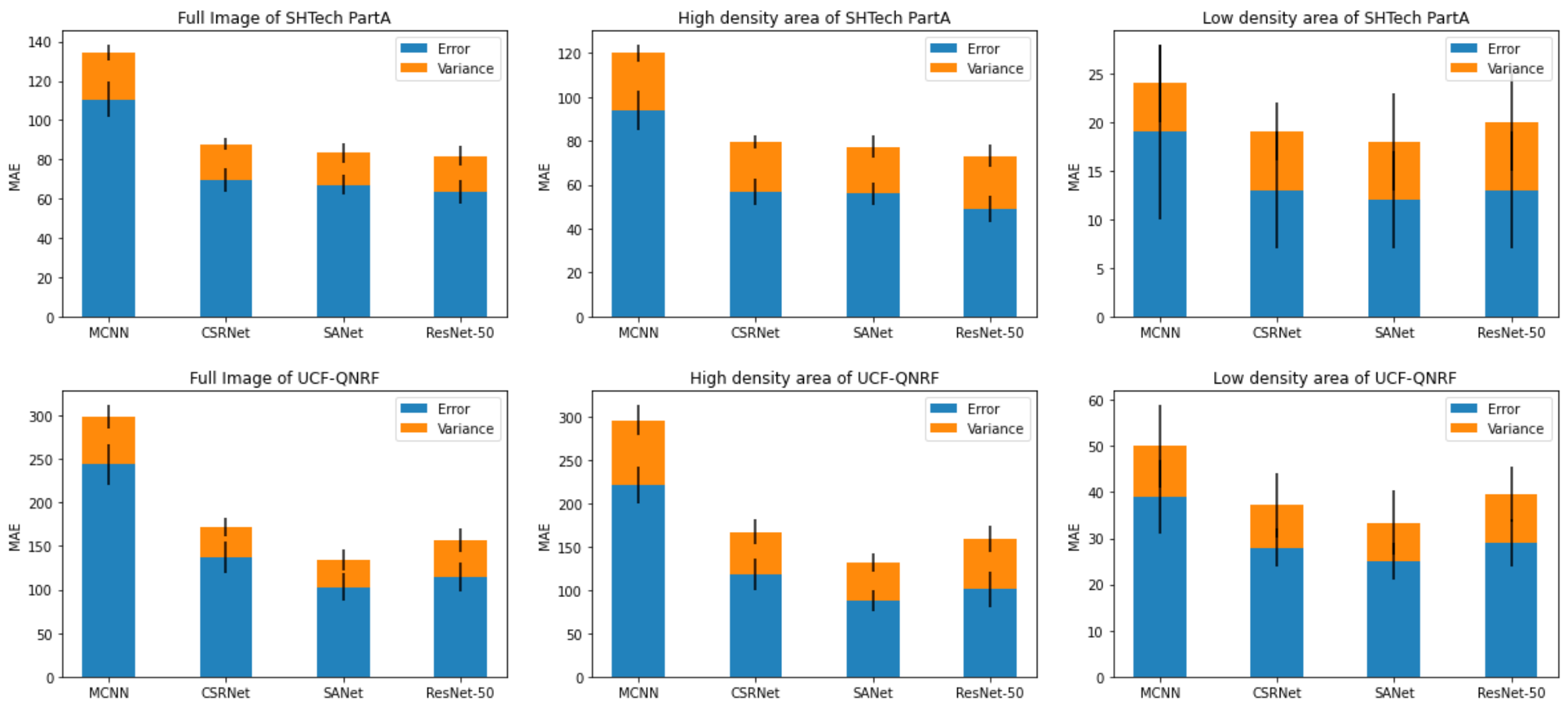}
   \vspace{-5mm}
   \caption{\small 
Comparative analysis of the prediction variance.
The variance refers to the difference in the results at different convergence states.
The error refers to the difference between the prediction and groundtruth.
Left to right are the analysis results of the full image, high-density area, and low-density area.
The results clearly show that there is a huge variance in prediction results.}
\label{fig:nosiy-analysis}
\end{figure}

\textbf{Ignorance of position and shape.}
To further clarify the large prediction variance, we visualized some examples.
Figure~\ref{fig:noisy-visulation} shows the obvious difference between the predicted density maps and the true position of the object (indicated by the red dot).
In some low-density areas, the prediction results ignore many objects (i.e., the density map does not cover many red dots).
Likewise, in some high-density regions, the crowd is poorly estimated (that is, the clustering on the density map is inconsistent with the trend of the red dots).
To sum up, these visualizations show that blindly improving spatial invariance does not learn the location and shape of objects.

\begin{table}[!t]
\vspace{-1mm}
\centering
\footnotesize
\caption{\small Cost on MCNN~(batchsize 1, image size 256$^2$).~LRA and TI refer to Low-Rank Approximation and Translation Invariance.~The Vanilla setting uses 256 Gaussian kernels per layer.}
\vspace{-1mm}
\footnotesize
\begin{tabular}{l|c|c|c}
\hline
Time (milliseconds) & Vanilla & LRA  & LRA+TI \\ \hline
Forward  &  51.3 &13.3 &  4.1   \\ \hline
Backward & 160.0 &  44.1&  12.6 \\ \hline
\end{tabular}
\label{tab:time}
\vspace{-6mm}
\end{table}

\begin{figure}[ht]
\vspace{-1mm}
\small
  \centering
   \includegraphics[width=0.95\linewidth]{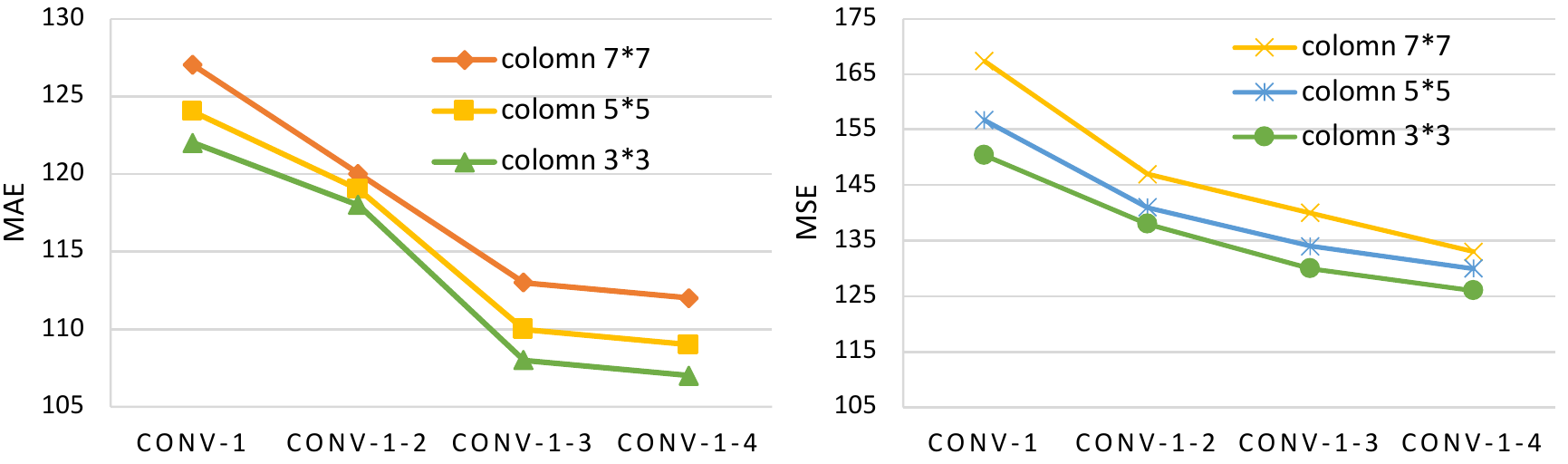}
   \vspace{-2mm}
   \caption{\small
The ablation study of MCNN\cite{MCNN-zhang}. 
The numbers after CONV indicate the range of usage of our GauNet layer.
}
   \label{fig:ablation-positon}
   \vspace{-6mm}
\end{figure}

\begin{table*}[t]
\footnotesize
\centering
\vspace{-2mm}
\caption{\small 
Comparison with the state-of-the-art methods on SHTech-PartA\cite{SHTech-zhang}, UCF\_CC\_50\cite{UCF_CC_50-idrees}, UCF-QNRF\cite{UCF-QNRF-idrees} and JHU-CROWD++\cite{JHU-CROWD-sindagi} datasets.
The best results are shown in bold.
This also applies to the following tables.}
\vspace{-2mm}
\begin{tabular}{llcccccccc}
\toprule
\multirow{2}{*}{Methods} & \multirow{2}{*}{Venue} & \multicolumn{2}{c}{SHTech-PartA} & \multicolumn{2}{c}{UCF\_CC\_50} & \multicolumn{2}{c}{UCF-QNRF} & \multicolumn{2}{c}{JHU-CROWD++} \\
                         &                        & MAE            & MSE             & MAE            & MSE            & MAE           & MSE          & MAE            & MSE            \\
\midrule
ADSCNet\cite{ADSCNet-bai}                  & CVPR’20                & 55.4           & 97.7            & 198.4          & 267.3          & \textbf{71.3}          & \textbf{132.5}        & -               & -               \\
AMSNet\cite{AMSNet-hu}                   & ECCV’20                & 56.7           & 93.4            & 208.4          & 297.3          & 101.8         & 163.2        & -               & -               \\
MNA\cite{MNA-wan}                      & NeurIPS’20             & 61.9           & 99.6            & -               & -               & 85.8          & 150.6        & 67.7           & 258.5          \\
DM-Count\cite{DM-Count-wang}                 & NeurIPS’20             & 59.7           & 95.7            & 211.0          & 291.5          & 85.6          & 148.3        & 66.0           & 261.4          \\
GLoss\cite{GLoss-wan}                    & CVPR’21                & 61.3           & 95.4            & -               & -               & 84.3          & 147.5        & 59.9           & 259.5          \\
URC\cite{URC-xu}                      & ICCV'21                & 72.8           & 111.6           & 294.0          & 443.1          & 128.1         & 218.1        & 129.7          & 400.5          \\
SDNet\cite{SDNet-ma}                    & ICCV'21                & 55.0           & 92.7            & 197.5          & 264.1          & 80.7          & 146.3        & 59.3           & 248.9         \\
\midrule
MCNN\cite{MCNN-zhang}                   & CVPR'16              & 110.2            & 173.2             & 377.6           & 509.1                & 277.0             & 426.0             & 188.9             & 483.4     \\
CSRNet\cite{CSR-li}                     & CVPR'18              & 68.2            & 115.0             & 266.1           & 397.5                & 119.2             & 211.4             & 85.9             & 309.2      \\
SANet\cite{SANet-cao}                     & ECCV'18              & 67.0            & 104.5             & 258.4           & 334.9                & -             & -             & 91.1             & 320.4    \\
\midrule
GauNet (MCNN)     & Ours & 94.2 & 141.8 & 282.6 & 387.2 & 204.2 & 280.4 & 165.3 & 486.6 \\
GauNet (CSRNet)   & Ours & 61.2 & 97.8  & 215.4 & 296.4 & 84.2  & 152.4 & 69.4  & 262.4 \\
GauNet (SANet)    & Ours & 59.2 & 95.4  & 209.2 & 278.4 & 86.6  & 162.8 & 68.9  & 270.6 \\
GauNet (ResNet-50) & Ours & \textbf{54.8} & \textbf{89.1}  & \textbf{186.3} & \textbf{256.5} & 81.6  & 153.7 &\textbf{58.2}  & \textbf{245.1}\\
\bottomrule
\end{tabular}
\label{tab:crowd-result-1}
\vspace{-2mm}
\end{table*}

\begin{table}[t]
\footnotesize
\centering
\vspace{-2mm}
\caption{\small Results on SHTech-PartB\cite{SHTech-zhang} dataset.}
\vspace{-3mm}
\begin{tabular}{llcc}
\toprule
\multirow{2}{*}{Methods} & \multirow{2}{*}{Venue} & \multicolumn{2}{c}{SHTech-PartB} \\
                         &                        & MAE             & MSE            \\
\midrule
ADSCNet\cite{ADSCNet-bai}                  & CVPR’20                & 6.4             & 11.3           \\
AMSNet\cite{AMSNet-hu}                   & ECCV’20                & 6.7             & 10.2           \\
DM-Count\cite{DM-Count-wang}                 & NeurIPS’20             & 7.4             & 11.8           \\
GLoss\cite{GLoss-wan}                    & CVPR’21                & 7.3             & 11.7           \\
URC\cite{URC-xu}                      & ICCV'21                & 12.0            & 18.7           \\
\midrule
MCNN\cite{MCNN-zhang}                     & CVPR’16                & 26.4            & 41.3           \\
CSRNet\cite{CSR-li}                   & CVPR’18                & 10.6            & 16.0           \\
SANet\cite{SANet-cao}                    & ECCV’18                & 8.4             & 13.2           \\
\midrule
GauNet (MCNN)              &    Ours                  & 17.6            & 24.7           \\
GauNet (CSRNet)            &    Ours                 & {7.6}           & {12.7}        \\
GauNet (SANet)            &     Ours                & 7.1             & 11.2           \\
GauNet (ResNet-50)          &     Ours              & \textbf{6.2}             & \textbf{9.9}           \\
\bottomrule
\end{tabular}
\label{tab:crowd-result-2}
\vspace{-3mm}
\end{table}

\begin{table}[t]
\footnotesize
\centering
\caption{\small Results on TRANCOS\cite{TRANCOS-guerrero} and MTC\cite{MTC-lu} dataset.}
\vspace{-3mm}
\begin{tabular}{lcccc}
\toprule
\multirow{2}{*}{Methods} & \multicolumn{2}{c}{TRANCOS} & \multicolumn{2}{c}{MTC} \\
                         & MAE          & MSE          & MAE        & MSE        \\
\midrule
ADMG\cite{ADMG-wan}        & 2.6          & 3.89         & -           & -           \\
TasselNetv2\cite{TasselNetv2-xiong} & -             & -             & 5.4        & 8.8        \\
S-DCNet\cite{S-DCNet-xiong}     & -             & -             & 5.6        & 9.1       \\
\midrule
CSRNet\cite{CSR-li}      & 3.56         & -             & 9.4        & 14.4       \\
GauNet (CSRNet) & {2.2}          & \textbf{2.6}          & {3.2}        & 4.6        \\
\midrule
GauNet (MCNN)   & 7.7          & 7.4          & 8.7        & 12.3        \\
GauNet (SANet)  & 2.5          & {2.8}          & 3.4        & {4.5}        \\
GauNet (ResNet-50) & \textbf{2.1}          & \textbf{2.6}          & \textbf{3.1}        & \textbf{4.3}  \\
\bottomrule
\end{tabular}
\label{tab:object-count-result}
\vspace{-5mm}
\end{table}

\vspace{-1mm}
\subsection{Ablation study}
\label{sec:ablation}
\vspace{-1mm}
We perform ablation studies with our method.
Due to space limitations, we only uses MCNN~\cite{MCNN-zhang} as an example.

\textbf{Effectiveness of accelerated modules}.
We conduct ablation studies to verify the effectiveness of low-rank approximation and translation invariance modules.
Table~\ref{tab:time} shows the experimental time cost of our proposed layer.
Compared with the original Gaussian convolution, our offered two acceleration modules can significantly improve the computational efficiency.

\textbf{Where should it be replaced?}
As shown in Figure~\ref{fig:ablation-positon}, we performed ablation studies on the three column convolutional structures of MCNN.
Overall, the three column structures have roughly the same results.
We noticed that replacing our layers in the first three convolutional layers will achieve larger improvements.
We also got similar results in other baselines.
Our method has fewer parameters than the original convolutional layer.
Thus in most cases, we replace all the convolutional layers (or 3$\times$3 convolutional layers in all residual blocks and pyramid pooling blocks) with our Gaussian convolutional layers.

\textbf{How to set the Gaussian kernels?}
Our method has three hyperparameters, i.e., the mean {\large $\mathbb{\mu}$}, variance $\mathbf{\sigma}$, and the number of Gaussian kernels $K$.
The mean value can be instantly set according to the stride of the original convolutional layer.
Thus we will only discuss variance $\mathbf{\Sigma}$ and the number of $K$ in the experiment.

As shown in Figure~\ref{fig:ablation-variance}, we have carried out studies on the three column structures of MCNN.
When the value of $K$ is large enough ($K=100$), we estimate the change of Gaussian kernel variance in each convolution layer.
We observe that the variance merely changed in the first convolutional layer.
Inspired by this, we usually set $K$ to $16$ with variance from $[-0.5,0.5]$ in the first two convolutional layers, and set $K$ to $2$ or $4$ in the successive convolutional layers with variance from $[-0.1,0.1]$).

\vspace{-1mm}
\subsection{Compare with state-of-the-art methods}
\label{sec:comparison}
\vspace{-1mm}
We compared our method with state-of-the-art methods in three applications~(crowd, vehicle, plant counting).

\textbf{Result of crowd counting}.~Table~\ref{tab:crowd-result-1} shows the results of crowd counting in the free camera perspective.
We took into account prediction variance and chose the average result for reporting.
Except for MCNN, the other three modified baselines outperform other state-of-the-art methods.
Compared to the original baselines, our variant has also achieved a huge improvement.
The performance of the light MCNN is even close to some of the most advanced methods.

Table~\ref{tab:crowd-result-2} shows the results in the surveillance scenarios.
Like free views, our model surpasses other state-of-the-art approaches, but the improvement in surveillance scenarios is not as much as free perspective.
We guess there is more noise in generating density maps in the free view.
Due to the noisy label in the groundtruth of SHTech-PartB, our method cannot further improve performance.

\textbf{Result of object counting}.~We also evaluated vehicle and plant counting.
Table~\ref{tab:object-count-result} shows that our model works well for vehicle scenarios.
The improvement is minor compared to the crowd counting because the vehicle scene holds less noise.
For plant counting, we got similar results.
Our model outperforms other state-of-the-art methods.
Notable is the improvement in the MSE metric, which shows that our method is more robust.
The overall performance is very close to the groundtruth.

\begin{figure*}[t]
\small
  \centering
   \includegraphics[width=0.81\linewidth]{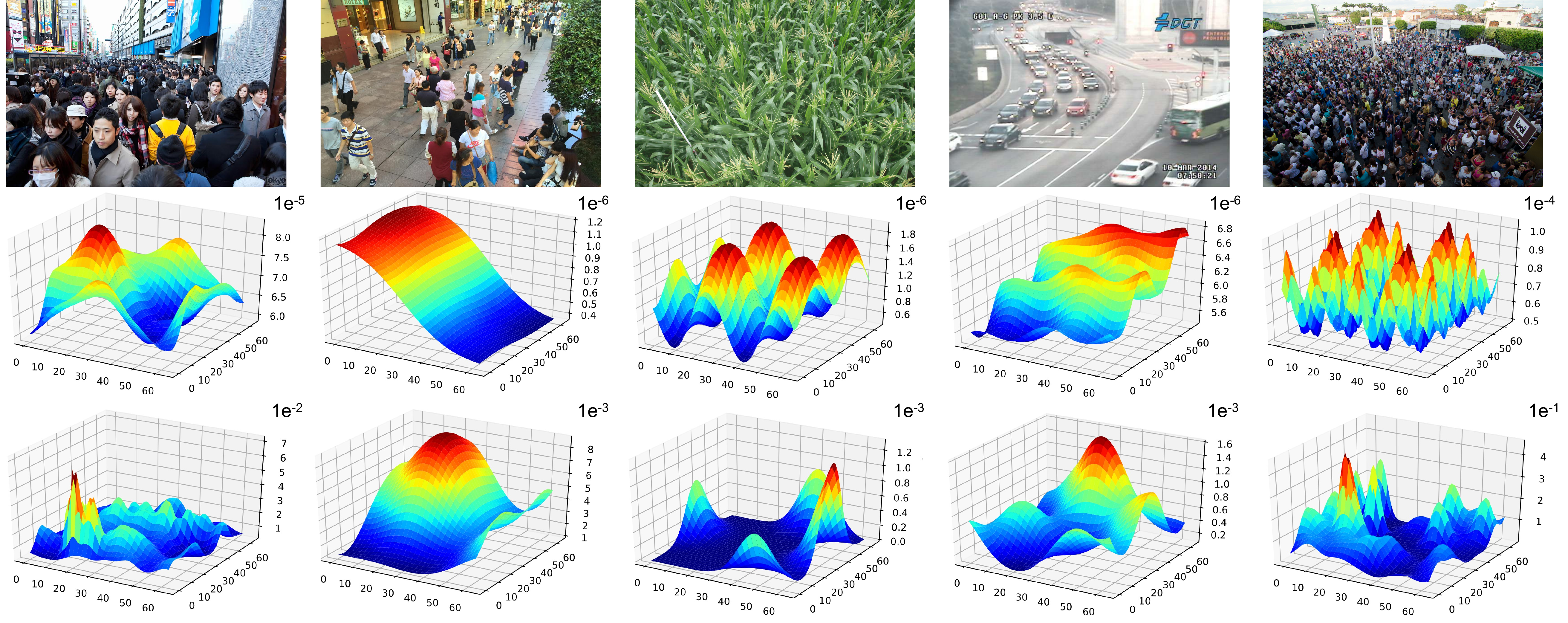}
   \vspace{-2mm}
   \caption{\small 
   The visualization of the convolution filters.
    From left to right are the results from SHTech-PartA, SHTech-PartB, MTC, TRANCOS and UCF-QNRF.
    From top to bottom are example images, our revised SANet and the original SANet\cite{SANet-cao}.
    Intuitively, our approach can fully understand the spatial information of objects and the perspective law of views.
    More details are shown in Sec. \ref{subsec:study}.
    }
   \label{fig:visualize-convolution}
   \vspace{-2mm}
\end{figure*}

\begin{table}[!t]
\vspace{-1mm}
\centering
\footnotesize
\caption{\small Robustness to annotation noise.~Both \cite{MNA-wan} and CSRNet adopt VGG backbone.~Results of VGG are from  Figure~4 of \cite{MNA-wan}.}
\vspace{-2mm}
\begin{tabular}{l|c|c|c|c|c}
\hline
MAE ($\downarrow$) & 0  & 4  & 8  & 16  & 32  \\
\hline
CSRNet (w/o)  & 119.2 & 125.4 & 133.7 & 142.5 & 166.2 \\
\hline
VGG~{\cite{MNA-wan}}  & 85.8  & 91.0  & 96.0  & 97.0  & 99.0  \\
\hline
CSRNet (ours) & 84.2  & 85.7  & 89.0  & 92.2  & 95.4 \\
\hline
\end{tabular}
\label{tab:noisy}
\vspace{-4.5mm}
\end{table}

\begin{figure*}[ht]
\small
  \centering
   \includegraphics[width=0.88\linewidth]{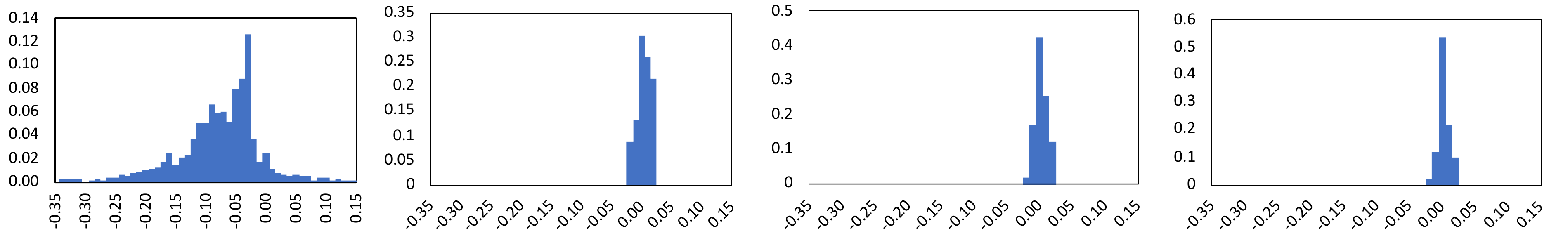}
   \vspace{-2.5mm}
   \caption{\small 
    Ablation study on MCNN\cite{MCNN-zhang}.
    From left to right are the variance optimization changes in the 1-st to 4-th convolutional layers.
    The abscissa indicates the range of change, 
    and the ordinate indicates the intensity of the change.}
   \label{fig:ablation-variance}
\vspace{-4mm}
\end{figure*}

\vspace{-1mm}
\subsection{Robustness to annotation noise}
\vspace{-1mm}
\label{sec:robustness}
We follow previous work \cite{MNA-wan} to verify robustness to annotation noise.
We generate a noisy dataset by randomly moving the annotation points by~\{4, 8, 16, 32\} pixels.
Then we train the model on noisy datasets with or without our proposed Gaussian convolutions.
Table~\ref{tab:noisy} shows the comparison.
Though the performances of all methods decrease as the annotation noise increases, our method is still more robust than other methods.
Figure~\ref{fig:noisy-visulation} also illustrates the predicted results of two examples with/without our method.

\vspace{-1mm}
\subsection{Visualization of convolution filters}
\label{subsec:study}
\vspace{-1mm}
We visualized the convolution filters to evaluate whether our model can simulate the density map generation and learn the spatial information of the objects.
Figure~\ref{fig:visualize-convolution} shows the results after visualization.
In general, our method can effectively learn the perspective law of the distribution of objects.
The results in the plant counting (column 3) are particularly obvious due to the more consistent scenarios.
Our method learns the planting distribution and even reflects the planting interval.
In contrast, the original SANet\cite{SANet-cao} only shows some noise in the image (e.g., marking Poles).
Similarly, our method also learns the distribution of pedestrians and vehicles by counting pedestrians and vehicles under the surveillance viewing angle (columns 2 and 4).
On the contrary, the original SANet blindly guesses high-density areas or overestimates low-density regions.
We also found similar results under the free perspective (columns 1 and 5), where our method can approximate crowd density distribution in pedestrian streets and squares.

\vspace{-1mm}
\section{Conclusion}
\label{sec:conclusion}
\vspace{-2mm}
We reveal the relationship between spatial invariance and density map noise.
Extensive experiments prove that if only instinctively improve the spatial invariance of CNNs, the model will easily overfit the density map noise. 
Inspired by this, we utilize a set of locally connected multivariate Gaussian kernels for replacing the convolution filters.
Unlike the pixelized-based filter, our proposed variant can approximately simulate the process of density map generation.
Considering the characteristics of object counting, we try to use translation invariance and low-rank approximation to improve efficiency.
Extensive experiments show that our method outperforms other state-of-the-art methods.
Our work points out the direction for future research.
It can avoid wildly improving the spatial invariance for the object counting. 
In the future, we will further analyze the relationship between the Gaussian kernel and spatial invariance.

\vspace{1mm}
{\footnotesize
\noindent \textbf{Acknowledgements.}~This work was partially supported by the Air Force Research Laboratory under agreement number~FA8750-19-2-0200;~the financial assistance award 60NANB17D156 from U.S. Department of Commerce, National Institute of Standards and Technology~(NIST);
the Intelligence Advanced Research Projects Activity~(IARPA) via Department of Interior/Interior Business Center~(DOI/IBC) contract number D17PC00340;
the Defense Advanced Research Projects Agency~(DARPA) grant funded under the GAILA program~(award HR00111990063).
The U.S. Government is authorized to reproduce and distribute reprints for Governmental purposes notwithstanding any copyright notation thereon.
The views and conclusions contained herein are those of the authors and should not be interpreted as necessarily representing the official policies or endorsements, either expressed or implied, of the Air Force Research Laboratory or the U.S. Government.}

\clearpage
\setcounter{section}{0}
\setcounter{figure}{0}
\setcounter{table}{0}
\renewcommand\thesection{\Alph{section}}
\renewcommand\thesubsection{\thesection.\arabic{subsection}}
\renewcommand\thefigure{A\arabic{figure}}

\section*{Supplementary materials}
\vspace{-1mm}
In supplementary material, we introduce the network structures and training details of all baselines.
\vspace{-2mm}
\section{Implementation details}
\label{sec:implementation details}
\vspace{-1mm}
Let's first present the network architectures of all baselines to facilitate understanding of the details.
\vspace{-1mm}
\subsection{Network structure of baselines}
\vspace{-1mm}
\noindent \textbf{MCNN baseline}.~We modified the original MCNN~\cite{MCNN-zhang} network by replacing the convolution filters in the first to fourth layers with locally connected low-rank Gaussian kernels.
As shown in Sec.~\ref{sec:ablation}, in the first two convolutional layers of three-column structures, each convolution filter is replaced with the equivalent 16 Gaussian kernels.
The Gaussian kernel variance of each dimension is fixed in the range of $[-0.5,0.5]$ for sampling.
Resembling the three-column convolution structure in the original MCNN (i.e., $3 \times 3$, $5 \times 5$, $7 \times 7$ branches), we correspondingly implemented three convolution structures with different Gaussian kernel numbers (i.e., maximum, middle, smaller column).
Specifically, the kernel number of the last two layers of the three branch structures (i.e., smaller, middle, and maximum) is set to $2$, $4$, and $6$, respectively.
At the end of the networks, we tallied a spatial pyramid pooling block to fuse the features of the three-column convolutional networks.
The structure of the spatial pyramid pooling block is shown in Figure \ref{fig:network architecture}.
Except for the first two convolutional layers, the variance of the remaining convolutional layer is fixed in the range of $[-0.1,+0.1]$ for sampling. 
The number of low-rank Gaussian kernels in the spatial pyramid pooling module is $8$.
The mean of the Gaussian kernel is selected according to the four times the stride of the original convolutional network.
Note that we also used the same settings in other baselines. 
So we will not discuss the mean of the Gaussian kernel later.

\noindent\textbf{CSRNet baseline}.~We retained the first ten convolutional layers of the VGG-16~\cite{vgg-14} backbone at the front end of CSRNet~\cite{CSR-li} network, and only modified the subsequent four dilated convolution branch structures.
The kernel number of the original A, B, C, and D branches are set to $2$, $4$, $6$, and $8$, respectively.
Note that the variance of every convolutional layer is fixed in the range of $[-0.1,+0.1]$ for sampling.
Unlike the original CSRNet, after comparing the performance of all branch structures, only the B branch with the stride length of $2$ was selected.
We applied a spatial pyramid pooling block on top of CSRNet to fuse the convolutional features of the four branches.
Similar to MCNN baseline, the number of low-rank Gaussian kernels in the spatial pyramid pooling module is also set as $8$.
According to the results of the ablation study on MCNN, since our modified layers are located at the back end of the original VGG-16 network, we did not use a large kernel number here.
In addition, our ablation experiment in CSRNet also proved the branch with the kernel number of $4$ achieves the best results.

\noindent \textbf{SANet baseline}.~We modified the first four-block convolutional layers in the original SANet~\cite{SANet-cao} network, and retained the deconvolutional layer at the end of the network.
The kernel number in the first two layers of the convolutional network is set to $16$.
While kernel number of the $1 \times 1$, $3\times3$, $5\times5$ and $7\times7$ convolution kernels in the latter two layers are set to $8$, $6$, $4$, and $2$, respectively.
Particularly, the range of the single dimension of the Gaussian kernel in the first two convolutional layers is $[-0.5,+0.5]$, while the range of the latter two layers is $[-0.1,+0.1]$.

\noindent \textbf{ResNet-50 baseline}.~$C^3$ framework~\cite{gao2019c} modified the original ResNet-50~\cite{ResNet-he} network and applied it to the crowd counting task.
Here we also used a similar setting.
We retain the first convolutional layer in the original ResNet-50 network.
Then we replace the $3 \times 3$ convolution filters in all residual blocks with locally connected low-rank Gaussian kernels.
The kernel number in all replacements is set to $4$.
Because of technical limitations, we correspondingly keep the $7 \times 7$ filter in the first convolution layer and $1 \times 1$ filters in the bottleneck layers.
To preserve the scale of the final density maps, we change the stride of the $3$rd convolutional layer from $2$ to $1$ as the encoder, and the decoder is composed of two convolutional layers.
We also implemented down-sampling with max-pooling instead of using convolutions with a stride of $4$.
The value range of the single dimension of the Gaussian kernel in all replacements is $[-0.1,+0.1]$.

\vspace{-1mm}
\subsection{Data preprocessing and training details}
\vspace{-1mm}
Following understanding the network structure, let's introduce the details of data processing and training.

\noindent \textbf{Data preprocessing}.~We carried out the preprocessing of input size and label transformation on all seven object counting datasets.
Specifically, we follow the setting of $C^3$ Framework~\cite{gao2019c}\footnote{\url{https://github.com/gjy3035/C-3-Framework}} to preprocess the SHTech-PartA~\cite{SHTech-zhang} and PatB~\cite{SHTech-zhang}, UCF-QNRF~\cite{UCF-QNRF-idrees} datasets in the crowd counting.
Meanwhile, the preprocessing steps of the remaining datasets are set according to the code repositories released by~\cite{KDMG-wan} to facilitate performance comparison.

\noindent \textbf{Training details}.~In addition to modifying some convolutional layers, the loss function and optimization process are set according to the original baseline.
Typically, we utilize the same training settings as the $C^3$ framework~\cite{gao2019c} and the previous work~\cite{KDMG-wan}.
Different from the original MCNN, CSRNet, and SANet implementations, we use the Batch-size training technique proposed by the $C^3$ framework to accelerate the training.
For MCNN and SANet baselines, except for the modified convolutional layer, the parameters of other parts are randomly initialized by a Gaussian distribution with a mean of $0$ and standard deviation of $0.01$.
Adam optimizer~\cite{adam-14} with a learning rate of $1\mathrm{e}{-5}$ is used to train the model.
For CSRNet, the first ten convolutional layers are from pre-trained VGG-16.
The other layers are initialized in the same way as MCNN.
Stochastic gradient descent (SGD) with a fixed learning rate of $1\mathrm{e}{-6}$ is applied during the training.
The revised ResNet-50 baselines are trained by stochastic gradient descent first.
Specifically, we employ the original ResNet-50 hyperparameters to pre-train on the ImageNet dataset, i.e., the learning rate of $0.1$, the momentum of $0.9$, weight decay of $1\mathrm{e}{-4}$, and a batch size of $256$.
Learning rate is reduced four times by a factor of $10$ at the $30$th, $60$th, $80$th, and $90$th epoch.
For the fine-tuning of ResNet-50, we adopt the same settings as the third-party code library $C^3$ Framework, i.e., the learning rate of $1\mathrm{e}{-4}$, weight decay is $0.995$, and the learning rate is reduced layer by layer.

\vspace{-1mm}
\section{More experimental results}
\vspace{-1mm}
In this section, we present more experiments to reveal the problem of the object counting task and prove the effectiveness of our method.
\vspace{-1mm}
\subsection{More ablation studies}
\vspace{-1mm}
The settings discovered in the ablation study on MCNN were directly applied to other baselines.
In addition, we also explore the setting of Gaussian kernel $K$ in all baselines.
In CSRNet, because we continue to use the first ten convolutional layers of the VGG-16 network, we can only apply our method at the back end of the original network.
The experimental results show that the branch structure with the $K$ value of $4$ achieved the best results.
However, the performance is still improved after the fusion of the entire four branch structures.
Therefore, we merge all four branches at the back end of CSRNet network, and set the $K$ value as $2$ to $32$ in each column structure.
We also verified the $K$ value in the residual block, and noticed that when the $K$ value is 4, the efficiency and performance have achieved the best balance.
In the spatial pyramid pooling of MCNN and SANet, we observed that the performance would increase and decrease as the $K$ value increased. 
We guess that this is caused by overfitting when the value of $K$ is too large. Therefore, the $K$ value in the spatial pyramid pooling layer is set as $16$.
\vspace{-1mm}
\subsection{Reveal the problem of object counting}
\begin{figure*}[t]
\small
 \centering
     \begin{subfigure}[b]{\textwidth}
          \centering
          \includegraphics[width=\linewidth]{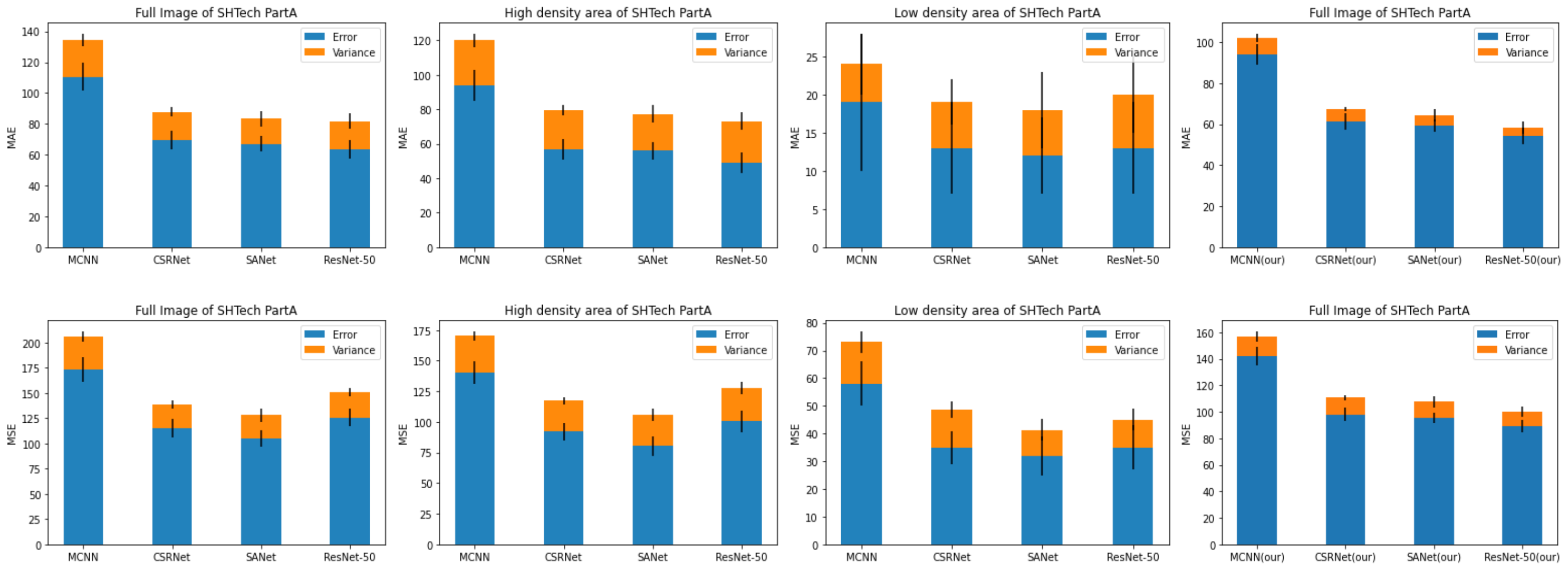}
          \caption{\footnotesize Comparative analysis of prediction variance and error on SHTech ParA~\cite{SHTech-zhang} dataset.}
    \end{subfigure}
    \begin{subfigure}[b]{\textwidth}
        \centering
        \includegraphics[width=\linewidth]{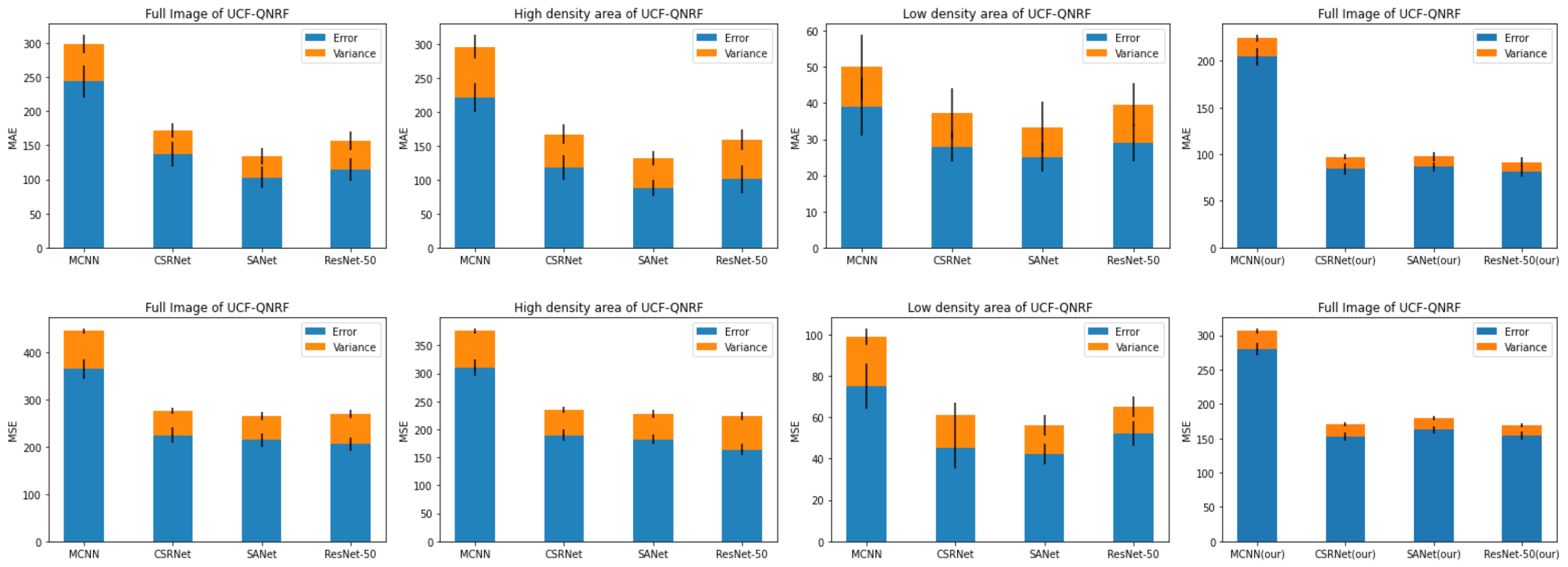}
        \caption{\footnotesize Comparative analysis of prediction variance and error on UCF-QNRF~\cite{UCF-QNRF-idrees} dataset.}
    \end{subfigure}
     \begin{subfigure}[b]{\textwidth}
        \centering
        \includegraphics[width=\linewidth]{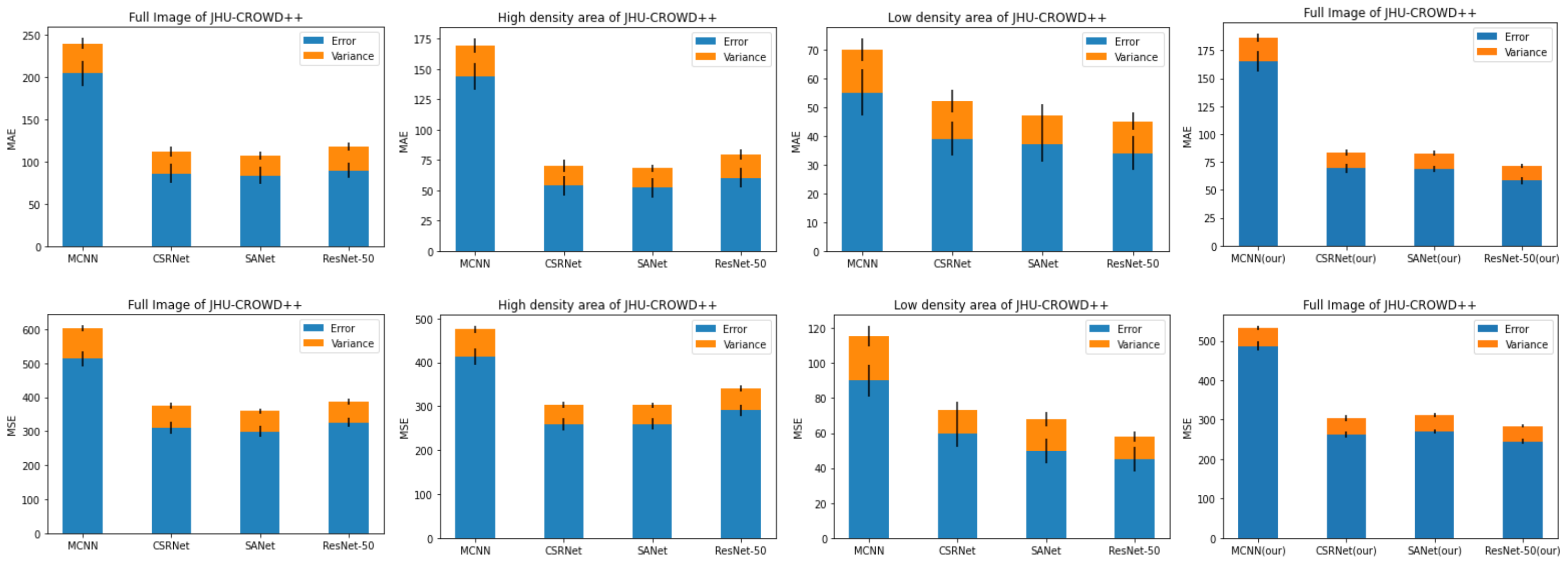}
        \caption{\footnotesize Comparative analysis of prediction variance and error on JHU-CROWD++~\cite{UCF-QNRF-idrees} dataset.}
    \end{subfigure}
   \vspace{-6mm}
   \caption{\small Comparative analysis of the variance and error of the prediction results.
The variance here refers to the difference in the prediction results for the same image at different convergence states. The error refers to the difference between the prediction and the ground truth. From left to right are the analysis results of the full image, high-density area, low-density area, and our modified baselines. The results clearly show that there is a huge variance in prediction results.~[\textbf{It is best to view in color and zoom in}].}
   \label{fig:nosiy-analysis-1}
   \vspace{-6mm}
\end{figure*}

\begin{figure*}[]
\small
   \centering
      \begin{subfigure}[b]{\textwidth}
          \centering
          \includegraphics[width=\linewidth]{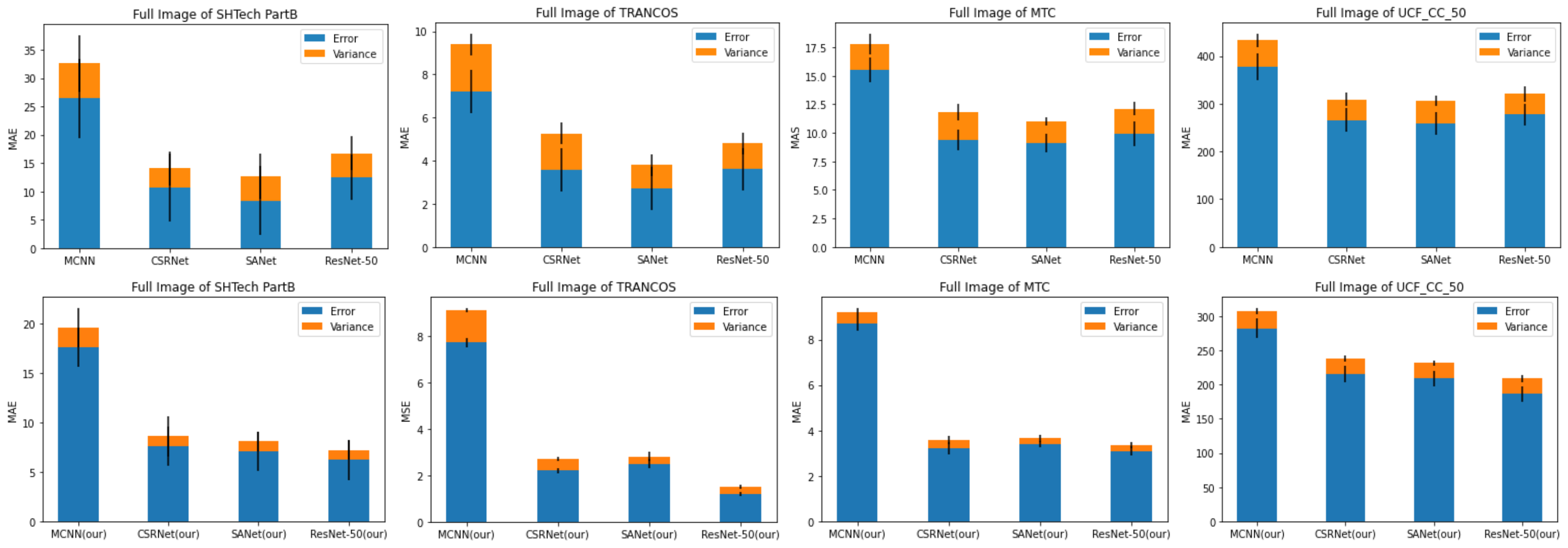}
          \caption{\footnotesize Comparative analysis of prediction variance and error on SHTech ParB~\cite{SHTech-zhang}, TRANCOS~\cite{TRANCOS-guerrero}, MTC~\cite{MTC-lu}, and UCF\_CC\_50~\cite{UCF_CC_50-idrees} datasets. Note that here are results on the MAE measurement.}
    \end{subfigure}
    \vspace{2mm}
    \begin{subfigure}[b]{\textwidth}
        \centering
        \includegraphics[width=\linewidth]{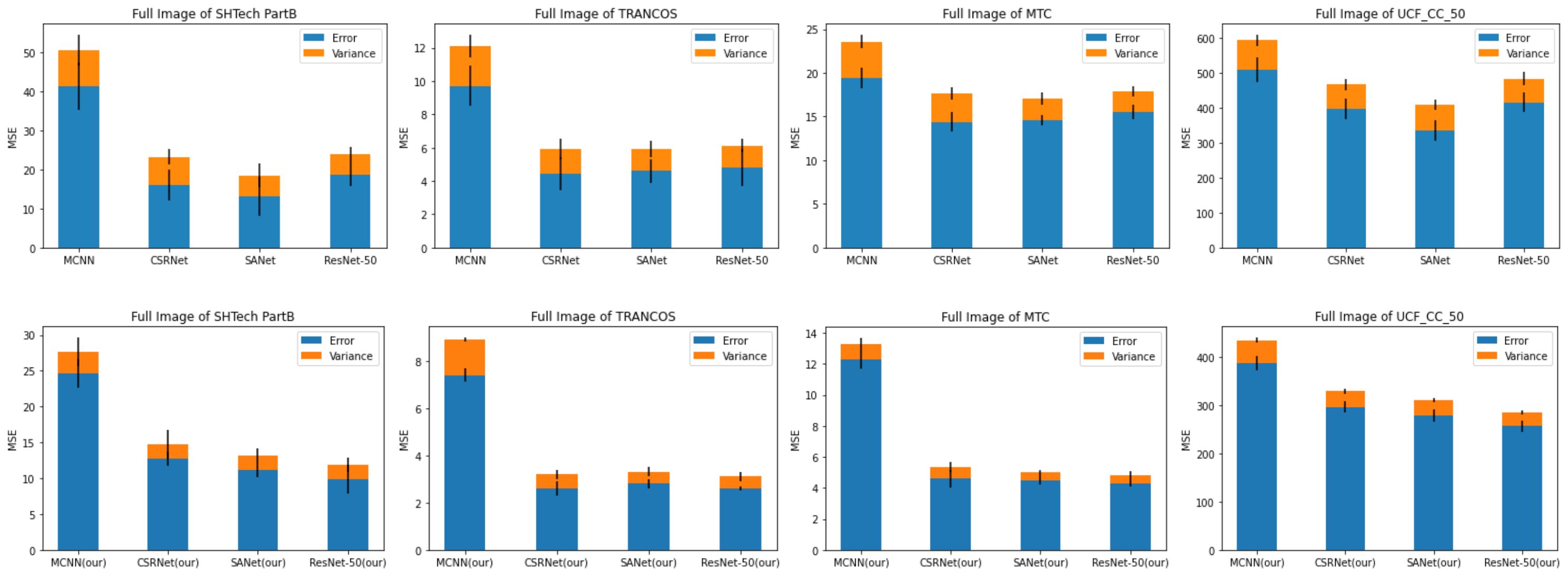}
           \caption{\footnotesize Comparative analysis of prediction variance and error on SHTech ParB~\cite{SHTech-zhang}, TRANCOS~\cite{TRANCOS-guerrero}, MTC~\cite{MTC-lu}, and UCF\_CC\_50~\cite{UCF_CC_50-idrees} datasets. Note that here are results on the MSE measurement.}
    \end{subfigure}
   \vspace{-8mm}
   \caption{\small The variance here refers to the difference in the prediction results for the same image at different convergence states. The error refers to the difference between the prediction and the ground truth. From left to right are the analysis results on SHTech ParB, TRANCOS, MTC, and UCF\_CC\_50 datasets. The results clearly show that there is a huge variance in prediction results.~[\textbf{It is best to view in color and zoom in}].}
   \label{fig:nosiy-analysis-2}
   \vspace{-4mm}
\end{figure*}

\vspace{-1mm}
As shown in Sec.~\ref{sec:reveal}, we illustrate more results in the supplementary material.
Here we test the performance of the different models that converge after randomly loading training data with the same hyperparameters.
To ensure the validity of the results, we repeated the training $20$ times in each baseline model.
Figures~\ref{fig:nosiy-analysis-1} and \ref{fig:nosiy-analysis-2} are the experimental results.
The variance we are discussing here refers to the prediction error of the trained model on the same image.
In other words, the variance shows that the model cannot converge stably.
Correspondingly, the error refers to the difference between the predicted result and the ground truth.

In general, we reach the same conclusions as in Sec.~\ref{sec:reveal}.
We distinguish high-density and low-density areas in SHTech ParA~\cite{SHTech-zhang}, UCF-QNRF~\cite{UCF-QNRF-idrees}, and JHU-CROWD++~\cite{JHU-CROWD-sindagi} to further analyze the results.
In these datasets, there are clear differences between high-density and low-density regions.
In general, the prediction variance of the original baseline is about a quarter of the total error.
This variance does not decrease as model performance (i.e. spatial invariance) increases.

In addition, we analyzed the full images in SHTech ParB~\cite{SSR-change}, TRANCOS~\cite{TRANCOS-guerrero}, MTC~\cite{MTC-lu} and UCF\_CC\_50~\cite {UCF_CC_50-idrees} dataset.
We no longer distinguish between high-density and low-density areas, because these datasets are relatively in low-density or high-density scenarios.
In general, we found that even in absolute low-density and high-density data sets, the model's prediction error is still very large.

We also analyzed the prediction variance of our modified baselines in Figures~\ref{fig:nosiy-analysis-1} and~\ref{fig:nosiy-analysis-2}. We found that our modified method significantly reduced the variance in the prediction results compared to the original network. 

{\small
\bibliographystyle{ieee_fullname}
\bibliography{egbibsu}
}

\end{document}